\documentclass[letterpaper, 10 pt, conference]{ieeeconf}  

\IEEEoverridecommandlockouts                              

\usepackage{graphicx}
\usepackage{multicol}
\usepackage{amsmath}

\usepackage{enumerate}
\usepackage{float}
\usepackage[caption=false]{subfig}
\floatstyle{plaintop}
\restylefloat{table}
\usepackage{multirow}
\usepackage{xfrac}
\usepackage[linesnumbered,lined,commentsnumbered,ruled,vlined]{algorithm2e}
\usepackage{siunitx}
\usepackage{caption} 
\let\labelindent\relax
\usepackage[shortlabels]{enumitem}
\usepackage{wrapfig}
\usepackage{bbm}
\usepackage{xr}
\usepackage{diagbox}
\usepackage[normalem]{ulem}
\usepackage{amsfonts}
\usepackage{xcolor}
\usepackage{mathtools}
\usepackage{comment}
\usepackage{microtype}

\makeatletter
\let\NAT@parse\undefined
\makeatother
\usepackage[pdfusetitle]{hyperref}
\hypersetup{pdfborder = 0 0 0,
  colorlinks = true,
  urlcolor = blue,
  linkcolor = black,
  citecolor = black,
}
\usepackage{cleveref}

\title{\LARGE \bf
ARCADE: Scalable Demonstration Collection and Generation via Augmented Reality for Imitation Learning
}

\author{Yue Yang$^{1}$, Bryce Ikeda$^{2}$, Gedas Bertasius$^{3}$, Daniel Szafir$^{4}$ \\ 
\href{https://yy-gx.github.io/ARCADE/}{ARCADE Homepage}
\thanks{$^{1}$Yue Yang, Department of Computer Science, the University of North Carolina at Chapel Hill, 232 S Columbia St, Chapel Hill, NC, USA
        {\tt\small yygx@cs.unc.edu}}%
\thanks{$^{2}$Bryce Ikeda, Department of Computer Science, the University of North Carolina at Chapel Hill, 232 S Columbia St, Chapel Hill, NC, USA
        {\tt\small bikeda@cs.unc.edu}}%
\thanks{$^{3}$Gedas Bertasius, Department of Computer Science, the University of North Carolina at Chapel Hill, 232 S Columbia St, Chapel Hill, NC, USA
        {\tt\small gedas@cs.unc.edu}}%
\thanks{$^{4}$Daniel Szafir, Department of Computer Science, the University of North Carolina at Chapel Hill, 232 S Columbia St, Chapel Hill, NC, USA
        {\tt\small daniel.szafir@cs.unc.edu}}%
}

\begin{document}

\maketitle
\thispagestyle{empty}
\pagestyle{empty}

\begin{abstract}


Robot Imitation Learning (IL) is a crucial technique in robot learning, where agents learn by mimicking human demonstrations. However, IL encounters scalability challenges stemming from both non-user-friendly demonstration collection methods and the extensive time required to amass a sufficient number of demonstrations for effective training. In response, we introduce the \textit{A}ugmented \textit{R}eality for \textit{C}ollection and gener\textit{A}tion of \textit{DE}monstrations (ARCADE) framework, designed to scale up demonstration collection for robot manipulation tasks. Our framework combines two key capabilities: 1) it leverages AR to make demonstration collection as simple as users performing daily tasks using their hands, and 2) it enables the automatic generation of additional synthetic demonstrations from a single human-derived demonstration, significantly reducing user effort and time. We assess ARCADE's performance on a real Fetch robot across three robotics tasks: \textit{3-Waypoints-Reach}, \textit{Push}, and \textit{Pick-And-Place}. Using our framework, we were able to rapidly train a policy using vanilla Behavioral Cloning (BC), a classic IL algorithm, which excelled across these three tasks. We also deploy ARCADE on a real household task, \textit{Pouring-Water}, achieving an 80\% success rate.

\end{abstract}


\section{Introduction}
\label{sec:intro}


Imitation Learning (IL) aims to empower end-users to teach robots skills and behaviors through demonstrations and has shown promising results in controlled laboratory environments~\cite{ravichandar2020recent, pan2020imitation, chen2021learning, chen2020joint}. Behavioral Cloning (BC)~\cite{pomerleau1988alvinn}, a common form of IL, mimics human actions from demonstrations using supervised learning, showing effectiveness in complex scenarios~\cite{nakanishi2004learning, liu2020understanding}. Compared to alternate approaches, such as adversarial imitation learning (AIL)~\cite{fu2018learning, ho2016generative}, BC stands out for its simplicity in implementation and optimization. Moreover, it functions offline, eliminating the need for potentially risky environmental interactions~\cite{hri2024, yang2022safe}. These advantageous features make BC a promising choice for deploying robots in real household environments. 
However, to realize this vision, we must overcome two significant BC challenges, the \textit{complex process of demonstration collection} and the \textit{data-hungry} nature of BC algorithms~\cite{george2023one}. 

Regarding the first challenge, current methods for gathering demonstrations often require familiarity with teleoperation using specific controllers (e.g., joystick, 3D mouse)~\cite{freymuth2023inferring, stepputtis2022system, sakr2020training} or contact-based kinesthetic teaching with robots~\cite{chen2020joint, hri2024}. Such approaches may not be feasible or desirable for non-expert users. Recently, Virtual Reality (VR) has been explored as a potential method to simplify demonstration collection process~\cite{george2023openvr, zhang2018deep}, but such approaches involve additional efforts such as creating realistic VR environments, leading to further complications. We are inspired by an alternative approach, suggested by \cite{duan2023ar2}, to use Augmented Reality (AR) to enable more natural collection of demonstrations. 

Regardless of the demonstration collection method, BC introduces a second challenge in requiring a substantial volume of expert demonstrations, often in the hundreds, for effective training. Amassing such a quantity of demonstrations may be excessively burdensome for end users. This challenge is primarily due to the covariate shift issue~\cite{ross2011reduction, chang2021mitigating}, where minor discrepancies in action prediction accumulate over time, leading the agent to encounter states not covered by the demonstrations. 



We introduce ARCADE, a novel AR-based framework tailored to effectively address both challenges. 
ARCADE provides a three-step process for generating demonstrations in a user-friendly and scalable manner. First, a user provides a single demonstration using their hands as they would in daily life, addressing the \textit{complex process of demonstration collection} challenge. During this process, the user wears an AR headset that tracks their hand motion and visualizes a robot digital twin overlaying the user. Second, ARCADE automatically generates a small set (10--15) of candidate demonstrations by following waypoints that are randomly sampled from the single user-collected demonstration, during which we apply a \textit{Key-Poses Detector} to preserve the core elements of the user's demonstration. These candidates are visualized in AR to the user who can rapidly filter out any that contain errors (e.g., violating user preferences or implicit constraints) to form a user-accepted set of demonstrations. Third, ARCADE generates a large set of additional high-quality demonstrations, all based on the initial user demonstration, and uses an \textit{Automatic Validation} approach to compare each candidate against the user's accepted set without further need for user input, thus rapidly scaling the demonstration set and addressing the \textit{data-hungry} challenge to enable effective BC training. We summarize our contributions as:


\begin{enumerate}[leftmargin=*, noitemsep]
    \item We introduce a novel framework 
    for generating demonstrations at scale from a single AR-captured demonstration. 
    \item Within this framework, we have developed two innovative techniques: a \textit{Key-Poses Detector} and \textit{Automatic Validation}, both designed to facilitate the generation of high-quality demonstrations from one user-provided AR demonstrations.
    \item We evaluate ARCADE on a physical Fetch robot for three manipulation tasks. The BC-trained policy with ARCADE-generated demonstrations demonstrates excellent performance across all tasks. Further validation of ARCADE in a more complex \textit{Pouring Water} task showed the robot achieving an 80\% success rate, highlighting ARCADE's potential for realistic robot assistance in homes.
\end{enumerate}

\section{Related Work}

\subsection{Demonstration Collection Methods}
Methods for collecting demonstrations in IL have evolved alongside the field itself. The earliest and also the most popular method involves users kinesthetically guiding the robot in a tactile manner to perform tasks~\cite{ravichandar2020recent}. However, the effectiveness of kinesthetic teaching depends on the user's dexterity and their ability to perform smooth demonstrations. This makes it challenging to gather large-scale datasets, as maintaining the stamina needed for repeated physical demonstrations is difficult. Teleoperation offers an alternative~\cite{goldberg1994beyond, lumelsky1993real, hokayem2006bilateral}, where users control the robot using various devices such as a keyboard and mouse~\cite{leeper2012strategies, kent2017comparison}, 3D-mouse~\cite{dragan2012online, stepputtis2022system, shridhar2023perceiver}, joysticks~\cite{laskey2017comparing}, or mobile phones~\cite{mandlekar2018roboturk} to perform tasks. 
Unfortunately, Jiang et al., 2024 shows that such systems typically require a longer training process for users to effectively operate these systems and they may achieve the worst performance compared to alternative methods~\cite{jiang2024comprehensive}. Recent research has explored the use of VR to streamline the process of demonstration collection, offering users a potentially more intuitive way to control robots~\cite{whitney2019comparing, zhang2018deep, lipton2017baxter, jaegle2021perceiver, george2023openvr, george2023one}. However, such VR methods necessitate the development of a simulation environment and remove the demonstration process from the contexts of real-world applications. To eliminate the need for a real robot during demonstration collection, Duan et al., 2023 suggests using AR for this purpose, which inspires our framework. 

\subsection{Behavioral Cloning}

Behavioral Cloning (BC), a key technique in Imitation Learning (IL), effectively completes tasks by replicating demonstrations provided by users~\cite{pomerleau1988alvinn}. For example, Ratliff et al., 2007  apply BC to manipulation and locomotion tasks by converting them into multiclass classification problems, subsequently solved via supervised learning~\cite{ratliff2007imitation}. However, one major challenge for BC methods is addressing covariate shift (also known as compounding errors)~\cite{ross2011reduction, chang2021mitigating}, which results in action predictions for out-of-distribution states. To address covariate shift, Dataset Aggregation (DAgger)~\cite{ross2011reduction} can improve robot policies during training, but may be taxing for users as it requires continuous feedback throughout the training process. An alternative solution is to supply a large number of high-quality demonstrations that encompass a broader state space, thereby enhancing the performance of the BC task~\cite{nakanishi2004learning, sasaki2018sample, reichlin2022back}, yet this too can be burdensome for users due to the stamina required to  create high quality large scale datasets. Recently, George et al., 2023 proposed generating multiple demonstrations from a single VR-collected demonstration~\cite{george2023one}. However, this approach does not guarantee alignment with user preferences, potentially affecting the quality of the generated demonstrations. Instead, our framework ensures all generated demonstrations match user preferences by deriving them from a user-approved set after reviewing AR-visualized options.

\section{Methodology}

\begin{figure*}[h]
    \centering
    \includegraphics[width=1.0\textwidth]{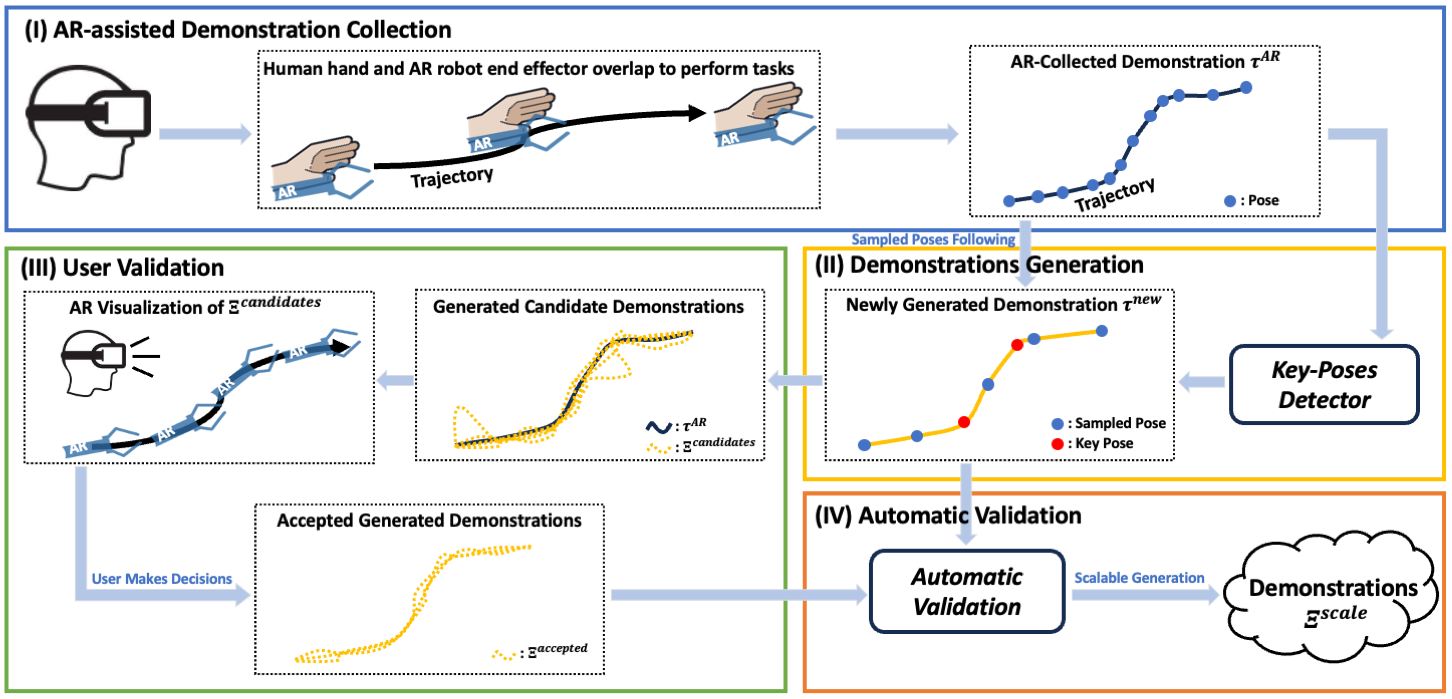}
    \caption{This figure shows the architecture of ARCADE. (I) First, a user provides a single demonstration, $\tau^{AR}$, through AR. (II) We generate a new demonstration, $\tau^{new}$, by following sampled poses, extracted from $\tau^{AR}$, and key poses, obtained via \textit{Key-Poses Detector}. (III) Additional candidate demonstrations are then generated, which are visualized in AR for user validation. Users filter the candidate demonstrations to form an accepted set of generated demonstrations, $\Xi^{accepted}$. (IV) Finally, we continue generating additional new demonstrations, automatically determining whether to keep or discard each demonstration based on comparing it to $\Xi^{accepted}$ via \textit{Automatic Validation}.
    }
    \label{fig:arch}
\end{figure*}

\subsection{Framework Overview}
\label{sucsec: framework_overview}

We introduce the ARCADE as a framework for generating demonstrations in a user-friendly and scalable manner, as illustrated in Figure~\ref{fig:arch}. 
\S\ref{subsec:demo_collection} details the initial AR-based user demonstration (Figure~\ref{fig:arch}A). Next, \S\ref{subsec:demo-generation} describes the method for generating demonstrations (Figure~\ref{fig:arch}B). Then, \S\ref{subsec:user_validation} details the method for users to validate generated demonstrations (Figure~\ref{fig:arch}C). Finally, \S\ref{subsec:auto_demo_gen} introduces an automated validation approach for these demonstrations to rapidly scale up the size of the demonstration set (Figure~\ref{fig:arch}D). 

Our framework integrates three underlying techniques: Markov Decision Processes (MDPs), Dynamic Time Warping (DTW), and Behavioral Cloning (BC). We model the environment using an MDP, denoted as $\mathcal{M}=\langle\mathcal{S}, \mathcal{A}, R, T, \rho_{0}\rangle$. Here, $\mathcal{S}$ represents the state space, for which we use the robot arm's joint values, and $\mathcal{A}$ denotes the action space, defined by changes in the arm's joint. The reward function is given by $R$, $T$ is the deterministic transition function, and $\rho_{0}$ represents the initial state probability distribution. We use DTW~\cite{muller2007dynamic}, an algorithm for quantifying the similarity between temporal sequences that may differ in timing or speed, as a crucial tool for \textit{Automatic Validation}. We employ BC as our IL algorithm, training a policy, $\pi_{\theta}$, to mimic demonstrations, $\Xi = \{\tau_i\}_{i=1}^M$, where each demonstration is a list of state-action pairs, $\mathcal{\tau} = \{(s_i, a_i)\}_{i=1}^{N}$.

\label{subsec:ar-demo-coll}

\subsection{AR-assisted Demonstration Collection}
\label{subsec:demo_collection}

\begin{figure}[H]
    \centering
    \includegraphics[width=1.0\columnwidth]{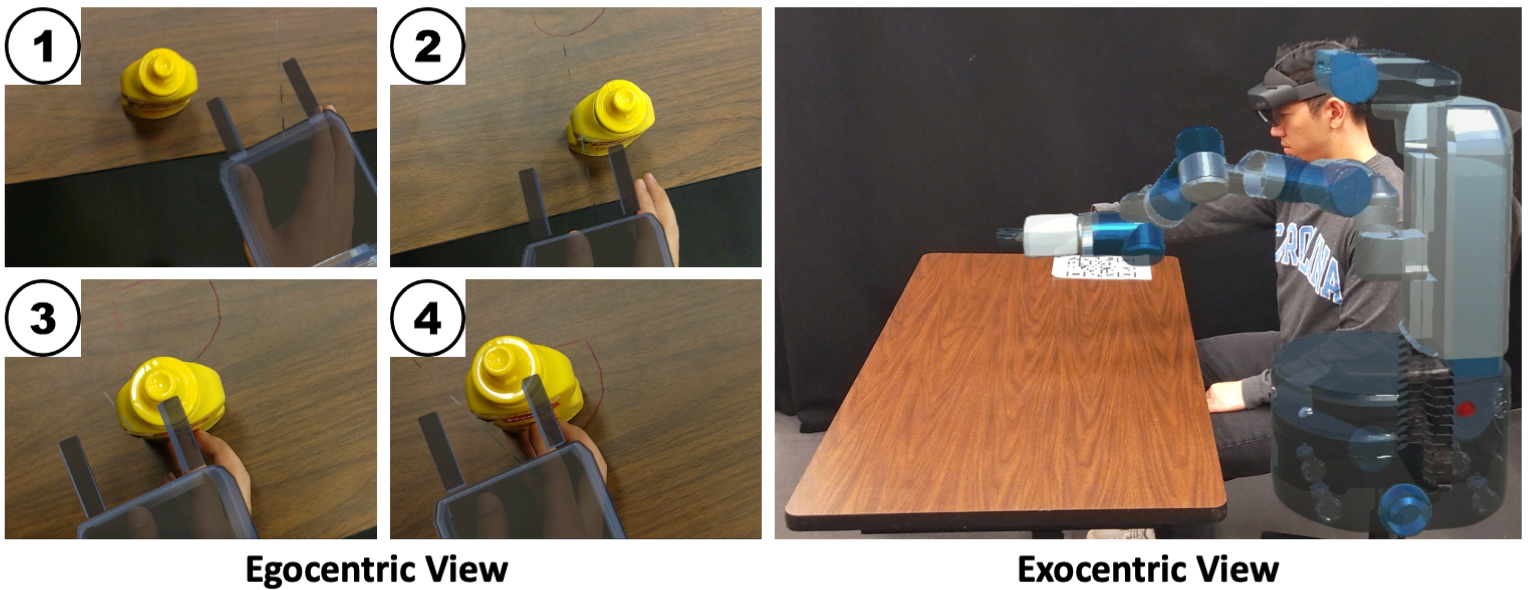}
    \caption{Left: Egocentric view showing the robot's end effector overlapping with and following the human hand's movements to perform the Push task. Right: Exocentric view showing how the human performs the task manually, with the digital twin robot end effector mirroring the hand movements.} 
    \label{fig:unity_env}
\end{figure}

To collect demonstrations of robot arm trajectories from users, we utilized the Microsoft HoloLens 2, an augmented reality head-mounted display (ARHMD). During the demonstration collection process (Figure \ref{fig:unity_env}), users wear the ARHMD, which overlays a digital twin of the robot on the user and provides an egocentric view of the robot's perspective. This setup facilitates real-time visual feedback of the robot's movements to the user. For our tasks and learning algorithm, the robot's end-effector must align with and track the user's hand. We accomplish this by using current inverse kinematics (IK) algorithms~\cite{rakita2018relaxedik}, which calculate the robot's joint angles based on the demonstrator's hand position. The distance between the user's pointer finger and thumb is used to inform gripper movements (open/close) for picking or placing objects. 
As the joint angles, end-effector positions, and pick or place actions are calculated during the demonstrations, this information is recorded on a separate machine, transmitted from the HoloLens via Unity Robotics Hub's ROS-TCP-Endpoint and ROS-TCP-Connector \cite{robotics_hub}.

With this setup, we record a single demonstration $\tau^{AR}$ from the user with a form as shown in Equation~\ref{eq:demo_form}. Each demonstration includes $N$ data points, with the form of end-effector pose, $p_i$, corresponding robot arm joints, $j_i$, and binary gripper state, $g_i$ for each timestep $i = 1,...,N$.

\begin{equation}
\label{eq:demo_form}
    \tau^{AR} \coloneqq \{(p_i, j_i, g_i)\}_{i=1}^{N}   
\end{equation}

\subsection{Demonstrations Generation}
\label{subsec:demo-generation}

Successfully collecting a single high-quality demonstration, $\tau^{AR}$, through user-friendly AR methods addresses one challenge, \textit{complex process of demonstration collection}, mentioned in \S\ref{sec:intro}. However, gathering a sufficient number of demonstrations with minimal user effort remains a hurdle. We automatically generate additional demonstrations based on the initial user-provided $\tau^{AR}$. Such generated demonstrations must meet two criteria for effective BC: 1) they should encompass a broader state space than the states in $\tau^{AR}$ and ensure task completion simultaneously, and 2) they must maintain similarity to $\tau^{AR}$, as IL algorithms are less effective with heterogeneous demonstrations~\cite{chen2021learning}.


To satisfy the first criterion, we use a waypoint-following approach. 
For waypoints sampling, we utilize a random interval length method to select poses from $\tau^{AR}$. Considering the poses, $\mathcal{P}^{AR} = \{p_i\}_{i=1}^{N}$, extracted from $\tau^{AR}$, we choose one pose at every $t$ timesteps, where $t$ is randomly determined within the range $[j, k]$ for each selection. By navigating through these randomly determined waypoints, the robot arm can explore a broader range of joints or end-effector states, irrespective of the specific definition of the state space.

 \begin{algorithm}[t]
    \caption{\textit{Key-Poses Detector}
    }
    \label{algo:keydatapoints_detector}
    \SetKwInOut{Input}{Input}\SetKwInOut{Output}{Output}
    \SetKwFunction{ComputeAngle}{ComputeAngle}
    \SetKwFunction{ComputeDensity}{ComputeDensity}
    \Input{\textit{points}: positions extracted from collected poses; \textit{window\_length}: the duration over which we compute pose angles and density; \textit{sharp\_turn\_threshold}; \textit{dense\_region\_threshold}.}


    $\textit{grasp\_release\_indices} \gets \{g_i^{grasp}\}_{i=1}^G \cup \{g_i^{release}\}_{i=1}^R$ \label{algo:keydatapoints_detector:line_gripper}\\

    $\textit{sharp\_turn\_indices}, \textit{dense\_region\_indices} \gets \text{Empty list}$ \\
    \For{\textit{idx}, \textit{point} \textbf{in} enumerate(\textit{points})}{
        $\textit{angle} \gets \ComputeAngle(\textit{point}, window\_length)$ \label{algo:keydatapoints_detector:line_sharpturn}\\
        \If{$\textit{angle} > \textit{sharp\_turn\_threshold}$}{ \label{algo:keydatapoints_detector:line_sharpturn_idx_1}
            $\textit{sharp\_turn\_indices}.append(idx)$ \label{algo:keydatapoints_detector:line_sharpturn_idx_2}
        }
        
        $\textit{density\_score} \gets \ComputeDensity(\textit{point}, window\_length)$ \label{algo:keydatapoints_detector:slow_movements}\\
        \If{$\textit{density\_score} > \textit{dense\_region\_threshold}$}{ \label{algo:keydatapoints_detector:slow_movements_idx_1}
            $\textit{dense\_region\_indices}.append(idx)$ \label{algo:keydatapoints_detector:slow_movements_idx_2}
        }
    }


    $\textit{key\_poses\_indices} \gets \textit{grasp\_release\_indices} \cup (\textit{sharp\_turn\_indices} \cap \textit{dense\_region\_indices})$ \\

    \Output{$\textit{key\_poses\_indices}$}
    
\end{algorithm}


However, to ensure critical waypoints necessary for task completion are not inadvertently filtered out during the sampling process, we introduce an automatic \textit{Key-Poses Detector}. Our method identifies key poses assuming that they occur either during grasping and releasing actions or at moments of significant angle changes in the user's hand trajectory coupled with slow movement (i.e., approaching zero velocity). Algorithm~\ref{algo:keydatapoints_detector} details the pseudocode for the \textit{Key-Poses Detector}. For grasp and release actions (Line~\ref{algo:keydatapoints_detector:line_gripper}), the collected demonstration, $\tau^{AR}$, inherently provides the required information. In situations involving angle changes, the function $ComputeAngle(\cdot)$ calculates the angle at the current position, using the start and end positions of the window to help identify sharp turns (Lines~\ref{algo:keydatapoints_detector:line_sharpturn_idx_1}-\ref{algo:keydatapoints_detector:line_sharpturn_idx_2}). Although the collected demonstration records only position-based data without velocity, we employ $ComputeDensity(\cdot)$ to gauge the density of neighboring poses within a window by calculating their average pairwise distances (Line~\ref{algo:keydatapoints_detector:slow_movements}), serving as a proxy to detect slow movements (Lines~\ref{algo:keydatapoints_detector:slow_movements_idx_1}-\ref{algo:keydatapoints_detector:slow_movements_idx_2}).

The combined set of sampled waypoints and detected key waypoints constitutes the waypoint set, denoted as $\mathcal{W}$. To reach these waypoints in $\mathcal{W}$, we use MoveIt's~\cite{coleman2014reducing} built-in motion planner~\cite{sucan2012open} and IK~\cite{kdl-url}, which tends to yield diverse trajectories, thereby encompassing a larger state space.

\subsection{User Validation}
\label{subsec:user_validation}

\begin{figure}[H]
    \centering
    \includegraphics[width=1\columnwidth]{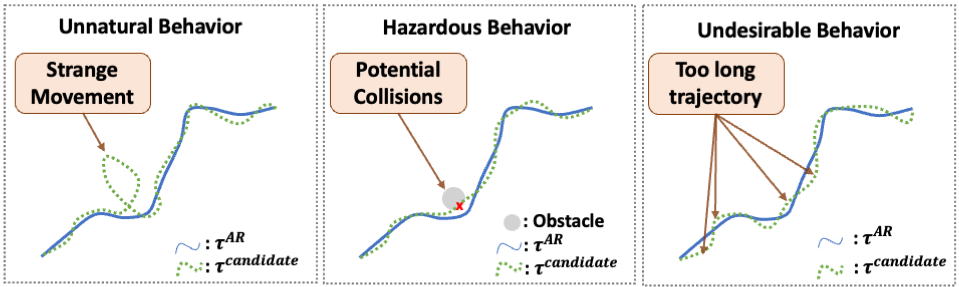}
    \caption{Visualized candidate demonstrations may exhibit behaviors that could lead to rejection by the user: (left) unnatural motions due to poor IK solutions, (middle) potentially hazardous motions, (right) misalignment with the user's preferences.}
    \label{fig:3_cases}
\end{figure}

Retaining key poses, as detailed in \S~\ref{subsec:demo-generation}, aids in meeting the second criterion: maintaining similarity between $\tau^{AR}$ and the newly generated demonstrations, $\tau^{new}$. However, this strategy alone is inadequate to tackle the problem of heterogeneous demonstrations fully because of three potential scenarios (illustrated in Figure~\ref{fig:3_cases}) that might occur in $\tau^{new}$: 1) unnatural movements resulting from IK instability~\cite{rakita2018relaxedik}; 2) potentially hazardous behaviors (e.g., the robot arm being too close to a table surface); and 3) any behaviors that the user may find undesirable (e.g., some users may favor shorter trajectories while some others might prioritize more human-like movements). Therefore, we must validate the behaviors in the generated demonstrations. To accomplish this, we initially create a set of generated candidate demonstrations, denoted as $\Xi^{candidates} \coloneqq \{\tau^{new}_i\}_{i=1}^{H}$. Subsequently, we present each candidate in the set $\Xi^{candidates}$ to the user for validation, via AR. Viewing demonstrations in AR enables users to identify any problems in $\tau^{new}$ and decide whether to retain or remove the demonstration. The result of this process is a set of user-accepted demonstrations, $ \Xi^{accepted} \coloneqq \{\tau^{accepted}_i\}_{i=1}^{H}$. We believe this interactive approach, where users observe and filter a small set of automatically generated demonstrations, may be substantially less demanding and more efficient for users compared to traditional methods in which users must manually generate their own additional demonstrations.

\subsection{Automatic Validation}
\label{subsec:auto_demo_gen}

 Effective training of BC often requires a large number of demonstrations, typically in the hundreds. Even with user's role shifted to validating generated demonstrations, the volume of necessary demonstrations for effective BC might still be daunting. Thus, we designed an automated method for scaling up the validation of generated demonstrations, where the user only needs to observe and approve a small set (e.g., 10--15 candidate demonstrations), after which the system can autonomously generate and self-validate candidates based on the characteristics of the user-approved set. 
 


 \begin{algorithm}[t]
    \caption{\textit{Automatic Validation}}
    \label{algo:auto_eval}
    \SetKwInOut{Input}{Input}\SetKwInOut{Output}{Output}
    \SetKwFunction{DTW}{DTW}
    \Input{\textit{$\tau^{new}$}: new candidate generated from $\tau^{AR}$; 
    \textit{$\Xi^{accepted}$}: accepted set of generated demonstrations;
    \textit{$\beta$}: the acceptable level.}
    
    $\Xi^{scale} \gets \text{Empty list}$ \\
    $\mathcal{S} \coloneqq \{\sum_{j=1, j\neq i}^H\DTW(\tau^{accepted}_i, \tau^{accepted}_j)\}_{i=1}^H$ \label{algo:auto_eval:line_S} \\
    $l \sim \text{Uniform}(1, H)$ \label{algo:auto_eval:line_delta_1} \\
    $\delta = \sum_{i=1, i\neq l}^H \DTW(\tau^{new}, \tau^{accepted}_i)$ \label{algo:auto_eval:line_delta_2} \\
    \If{$\delta \leq \beta\min(\mathcal{S})$}{  \label{algo:auto_eval:comp_1} 
        $\Xi^{scale}.append(\tau^{new})$  \label{algo:auto_eval:comp_2} 
    } 
    \Output{$\Xi^{scale}$} 
\end{algorithm}

As shown in Algorithm~\ref{algo:auto_eval}, we leverage $\Xi^{accepted}$ to construct a similarity array, $\mathcal{S}$, utilizing dynamic time warping, $DTW(\cdot)$ (\S\ref{sucsec: framework_overview}). Each element in $\mathcal{S}$ quantifies the similarity for every pair of user-accepted demonstrations, $\Xi^{accepted}$.
Subsequently, we assess each newly generated candidate demonstration, $\tau^{new}$, against $\Xi^{accepted}$, excluding one randomly for fair comparison, utilizing $DTW(\cdot)$ (line~\ref{algo:auto_eval:line_delta_1}-\ref{algo:auto_eval:line_delta_2}). Acceptance of the newly generated demonstration, $\tau^{new}$, only occurs when $\delta \leq \beta\min(\mathcal{S})$ (line~\ref{algo:auto_eval:comp_1}-\ref{algo:auto_eval:comp_2}). The parameter $\beta$, a scalar (e.g., 0.95), determines the acceptable level: a higher acceptable level (i.e., smaller $\beta$) results in more similar generated demonstrations but smaller coverage of the state space, and vice versa. This automatic filtering mechanism ensures that we only retain demonstrations aligning with user preferences. This process eliminates the need for constant user supervision (i.e., after the initial interaction where the user provides a single demonstration of their own and then filters generated candidates to the approved set of 10--15 demonstrations, no further user input is needed) and thus enhances the scalability of BC demonstration generation. Following the scalable, automated generation of demonstrations, BC models can be trained as usual, using the extensive set of demonstrations, $\Xi^{scale}$. 


\section{System Validation}



\begin{figure}[H]
    \centering
    \includegraphics[width=0.5\textwidth]{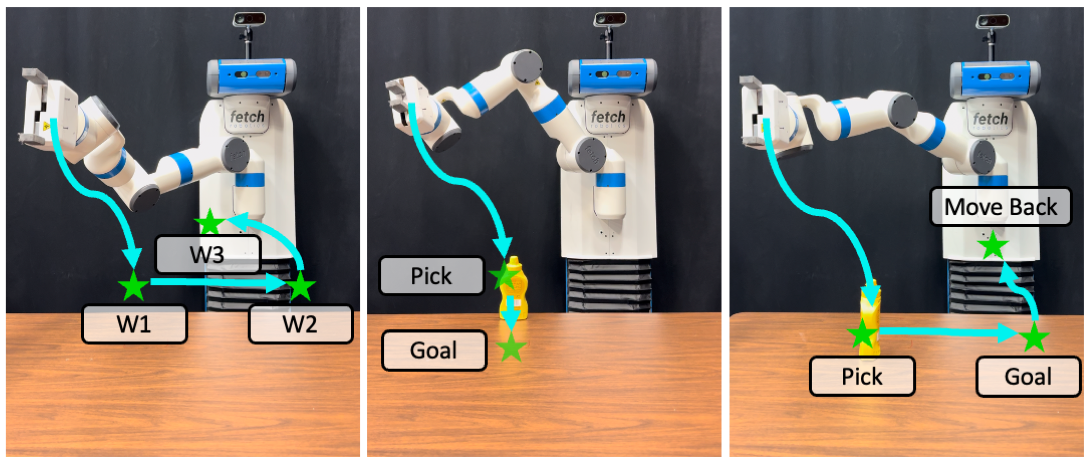}
    \caption{We evaluate on three tasks: Left: 3-Waypoints-Reach, Middle: Push, Right: Pick-And-Place.}
    \label{fig:tasks_description}
\end{figure}

We evaluate our framework using a real Fetch robot. We first examined performance on three archetypal tasks, \textit{3-Waypoints-Reach}, \textit{Push}, and \textit{Pick-And-Place} (Figure~\ref{fig:tasks_description}), chosen because they exemplify fundamental manipulation behaviors that, when combined, can accomplish a variety of complex household activities. We provide an example of this in a fourth, more complex \textit{Pouring Water} task (\S\ref{subsec:water}). 

In the \textit{3-Waypoints-Reach} task, the robot arm aims to hit three predefined waypoints: $W_1$, $W_2$, and $W_3$. For the \textit{Push} task, the robot arm must push an object on the desk from its starting position to a predefined goal point. The \textit{Pick-And-Place} task involves the robot arm grasping an object, moving it to another location, and then releasing it. For \textit{3-Waypoints-Reach} and \textit{Push}, the state space consists of 7 arm joints, and the action space includes 7 delta arm joints, corresponding to each joint's movement. For \textit{Pick-And-Place}, we expand the state space to eight dimensions to include the gripper angle, and the action space also increases to eight dimensions, encompassing the delta changes in the gripper.

We introduce a \textit{Task Completion Error (TCE)} metric, measured in meters, to evaluate BC performance across the three archetypal tasks. For the \textit{3-Waypoints-Reach} task, we calculate this metric by averaging the minimum distances to the three waypoints during the evaluation. For the \textit{Push} and \textit{Pick-And-Place} tasks, it measures the distance from the object's final position to its target goal. 

To benchmark ARCADE's effectiveness, we compared four BC policies across 3 archetypal tasks:
\begin{itemize}
    \item ARCADE ($\tau^{AR}$): a policy trained using just the initial user AR demonstration from ARCADE.
    \item ARCADE ($\Xi^{scale}$): a policy trained with the full ARCADE system, consisting of 100 generated demonstrations.
    \item BL ($\tau^{ki}$): a baseline (BL) policy trained using a single demonstration collected via kinesthetic teaching, $\tau^{ki}$.
    \item BL ($\Xi^{scale\_ki}$): a baseline policy trained using 100 demonstrations, $\Xi^{scale\_ki}$, generated identically to $\Xi^{scale}$ but based on $\tau^{ki}$ instead of $\tau^{AR}$.
\end{itemize}
To assess each task, we execute each of the four BC-learned policies ten times, reporting the mean and standard deviation of the \textit{TCE} for each policy as our results.


\subsection{Implementation Details}

\subsubsection{Hardware}
We utilized the Microsoft HoloLens 2 as our augmented reality head-mounted display (ARHMD). 
The HoloLens 2 tracks both the position and orientation of the wearer's hands, allowing us to map the user's hand movements to the robot during the initial demonstration. It also enables the visualization of virtual imagery, which we use to align a digital twin of the robot with the user's movements during the demonstration and to display candidate demonstrations for the user to review and select for the accepted demonstration set. 
Our current implementation utilizes the Fetch robot, a mobile manipulation platform with a 7 degree-of-freedom arm, although our work can be generalized to any robot manipulator. 

\subsubsection{Policy Architecture}
To represent the BC policy $\pi_{\theta}$, we use a GaussianMLP model~\cite{garage}. This model is composed of two multilayer perceptrons, one for producing the mean $\mu$ and the other for the standard deviation $\sigma$, together forming a Gaussian distribution. The robot arm's action is then sampled from this distribution, denoted as $a \sim \pi_{\theta}$.

\subsection{Archetypal Task Results}

\begin{figure}[H]
    \centering
    \includegraphics[width=0.5\textwidth]{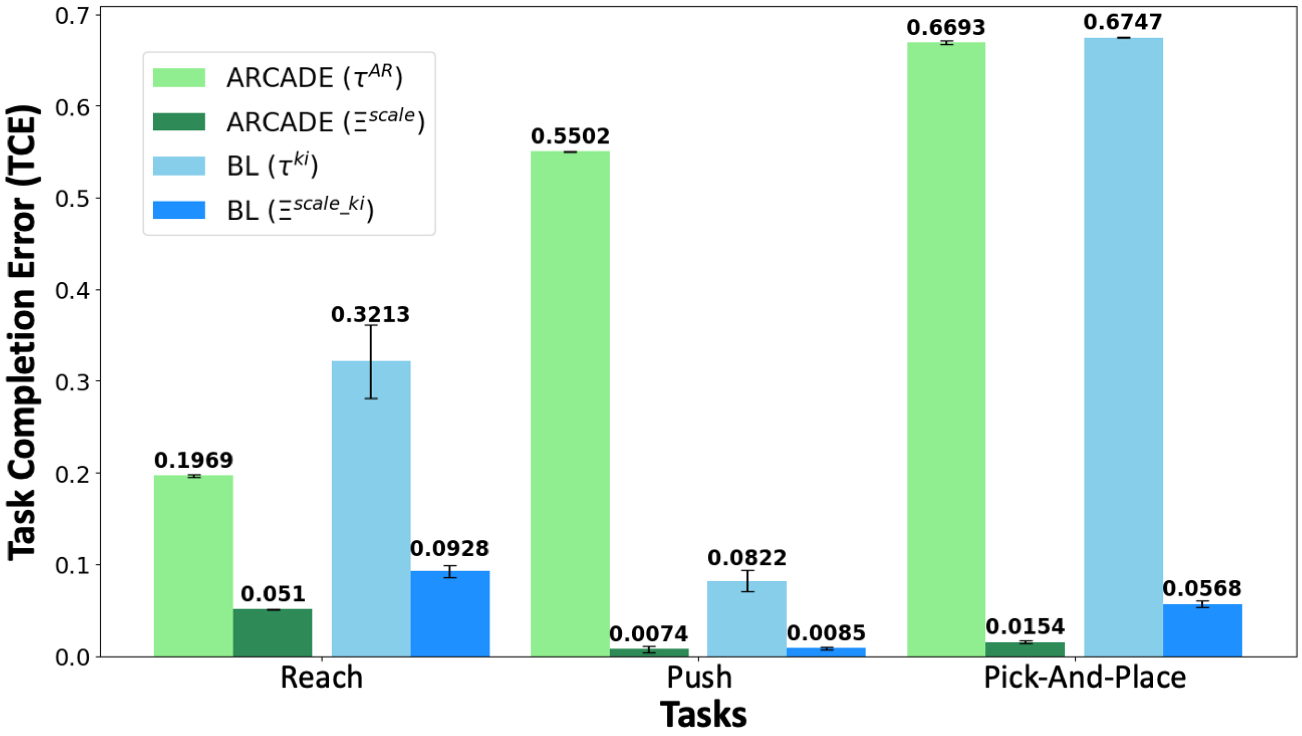}
    \caption{The results of BC policies trained using ARCADE or a kinesthetic teaching baseline (BL) with either 1 or 100 demonstrations across three tasks. The full (100 demonstration set: $\Xi^{scale}$) ARCADE system offers the best performance in all tasks.
    }
    \label{fig:bar_chart}
\end{figure}

\begin{figure*}[h]
    \centering
    \includegraphics[width=1.0\textwidth]{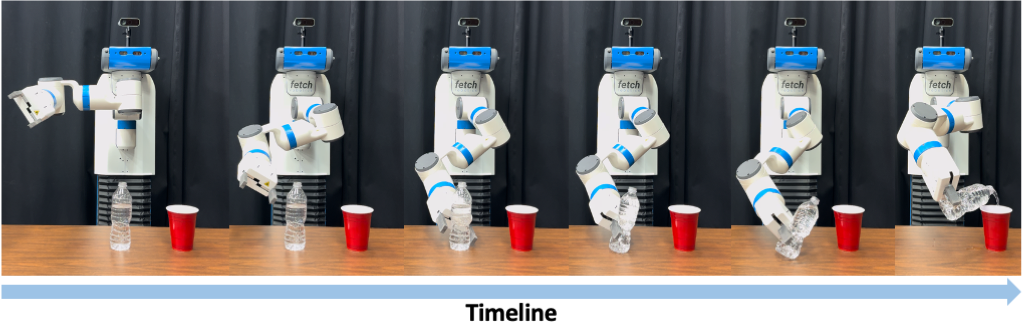}
    \caption{The robot successfully pours water from a single-user demonstration, showcasing ARCADE's effectiveness in equipping robots for practical household chores.
    }
    \label{fig:pour_water}
\end{figure*}

Figure~\ref{fig:bar_chart} illustrates the results of each of the four BC policies across the three archetypal tasks. We conducted a two-way analysis of variance (ANOVA) to test whether the type of demonstration collection (kinesthetic teaching baseline vs our ARCADE framework) and the size of the demonstration set ($|\Xi|=1$ or $|\Xi|=100$) affected BC policy TCE. We found significant main effects of both factors and their interaction on TCE at $p<.001$ for all three tasks. Using Tukey's Honestly Significant Difference (HSD) test to compare the performance of all four BC policies, we found that all four policies performed significantly differently ($p<.0001$ for each comparison), with performance ordered from best to worst as the {ARCADE ($\Xi^{scale}$)} (best), {BL ($\Xi^{scale\_ki}$)}, {ARCADE ($\tau^{AR}$)}, and {BL ($\tau^{ki}$)} (worst) for the {3-Waypoint-Reach} and {Pick-and-Place} tasks. For the push task, Tukey's HSD did not reveal a significant difference between the performance of the {ARCADE ($\Xi^{scale}$)} and {BL ($\Xi^{SCLAE\_ki}$)} ($p=.976)$, with both of them significantly better than the {BL ($\tau^{ki}$)} ($p<.0001$), which itself outperformed the {ARCADE ($\tau^{AR}$)} ($p<.0001$). These findings indicate that the ARCADE framework can generate demonstrations that match or surpass those from traditional kinesthetic teaching in terms of BC policy performance. Furthermore, the results show that both sets of demonstrations generated by our method, $\Xi^{scale}$ and $\Xi^{scale\_ki}$, facilitate effective BC training. 




\subsection{Real Household Task - Pouring Water}
\label{subsec:water}

To demonstrate ARCADE's capability in handling more complex household tasks and its potential for widespread home robot deployment, we introduce an additional task: \textit{Pouring-Water}. Here, the goal is for the robot to learn to grasp a bottle and pour water into a cup from just a single demonstration given by the user. This task utilizes the same state and action spaces as the \textit{Pick-And-Place} task and is deemed successful when water is poured into the cup. Testing the BC policy trained with $\Xi^{scale}$ from ARCADE, we achieved an 80\% success rate (8 of 10 trials), with failures attributed to the plastic bottle's shape alteration. Figure~\ref{fig:pour_water} captures a successful instance of the robot performing the pouring action.

\section{Conclusion}

We introduce ARCADE, a scalable framework that allows the collection of numerous high-quality demonstrations from a single user-collected demonstration via AR. This approach offers a user-friendly and time-efficient method for demonstration collection. Empirical evaluations across three archetypal robot tasks demonstrate ARCADE's effectiveness in generating high-quality demonstrations suitable for effectively training IL algorithms. Applying ARCADE to the real household task of \textit{Pouring-Water} illustrates the framework's potential to facilitate the widespread integration of robots into daily home life.


\section{Acknowledgements}
This work was sponsored by the National Science Foundation (NSF) under award \#2222953. This work was also supported by the Sony Faculty Innovation Award, Laboratory for Analytic Sciences via NC State University, ONR Award N00014-23-1-2356.

\bibliographystyle{IEEEtran}
\bibliography{refs}

\end{document}